\documentclass{MITcsail}
\usepackage{hyperref}

\usepackage{algorithm}
\usepackage{algpseudocode}

\usepackage{booktabs}
\newcommand{\ttx}[1]{\ttfamily{#1}}
\usepackage[symbol]{footmisc}
\usepackage{listings}
\usepackage{xcolor}
\usepackage{graphics}
\usepackage{wrapfig}
\usepackage{placeins}

\definecolor{codegreen}{rgb}{0,0.6,0}
\definecolor{codegray}{rgb}{0.5,0.5,0.5}
\definecolor{codepurple}{rgb}{0.58,0,0.82}
\definecolor{backcolour}{rgb}{0.95,0.95,0.92}

\lstdefinestyle{mystyle}{
    backgroundcolor=\color{backcolour},   
    commentstyle=\color{codegreen},
    keywordstyle=\color{magenta},
    numberstyle=\tiny\color{codegray},
    stringstyle=\color{codepurple},
    basicstyle=\ttfamily\footnotesize,
    breakatwhitespace=false,         
    breaklines=true,                 
    captionpos=b,                    
    keepspaces=true,                 
    numbers=left,                    
    numbersep=5pt,                  
    showspaces=false,                
    showstringspaces=false,
    showtabs=false,                  
    tabsize=2
}

\title{PyHopper - Hyperparameter optimization}

\author{Mathias Lechner$^{1,2,\dagger}$, Ramin Hasani$^{1,2}$, Philipp Neubauer$^{2,3}$, Sophie Neubauer$^{2,3}$, Daniela Rus$^1$ \\
\vspace{1em} 
\normalfont{\small $^{1}$Massachusetts Institute of Technology (MIT)}\\
\normalfont{\small $^{2}$Simple AI}\\
\normalfont{\small $^{3}$DatenVorspung GmbH}\\
\normalsize{\small $^{\dagger}$Correspondence E-mail: mlechner@mit.edu}\\
}

\begin{document}

\maketitle
\thispagestyle{firstpagestyle} 

\begin{abstract}
Hyperparameter tuning is a fundamental aspect of machine learning research. Setting up the infrastructure for systematic optimization of hyperparameters can take a significant amount of time.
Here, we present PyHopper, a black-box optimization platform designed to streamline the hyperparameter tuning workflow of machine learning researchers. 
PyHopper's goal is to integrate with existing code with minimal effort and run the optimization process with minimal necessary manual oversight.
With simplicity as the primary theme, PyHopper is powered by a single robust Markov-chain Monte-Carlo optimization algorithm that scales to millions of dimensions. Compared to existing tuning packages, focusing on a single algorithm frees the user from having to decide between several algorithms and makes PyHopper easily customizable.
PyHopper is publicly available under the Apache-2.0 license at \url{https://github.com/PyHopper/PyHopper}.
\begin{figure}[h]
    \centering
    \includegraphics[width=\textwidth]{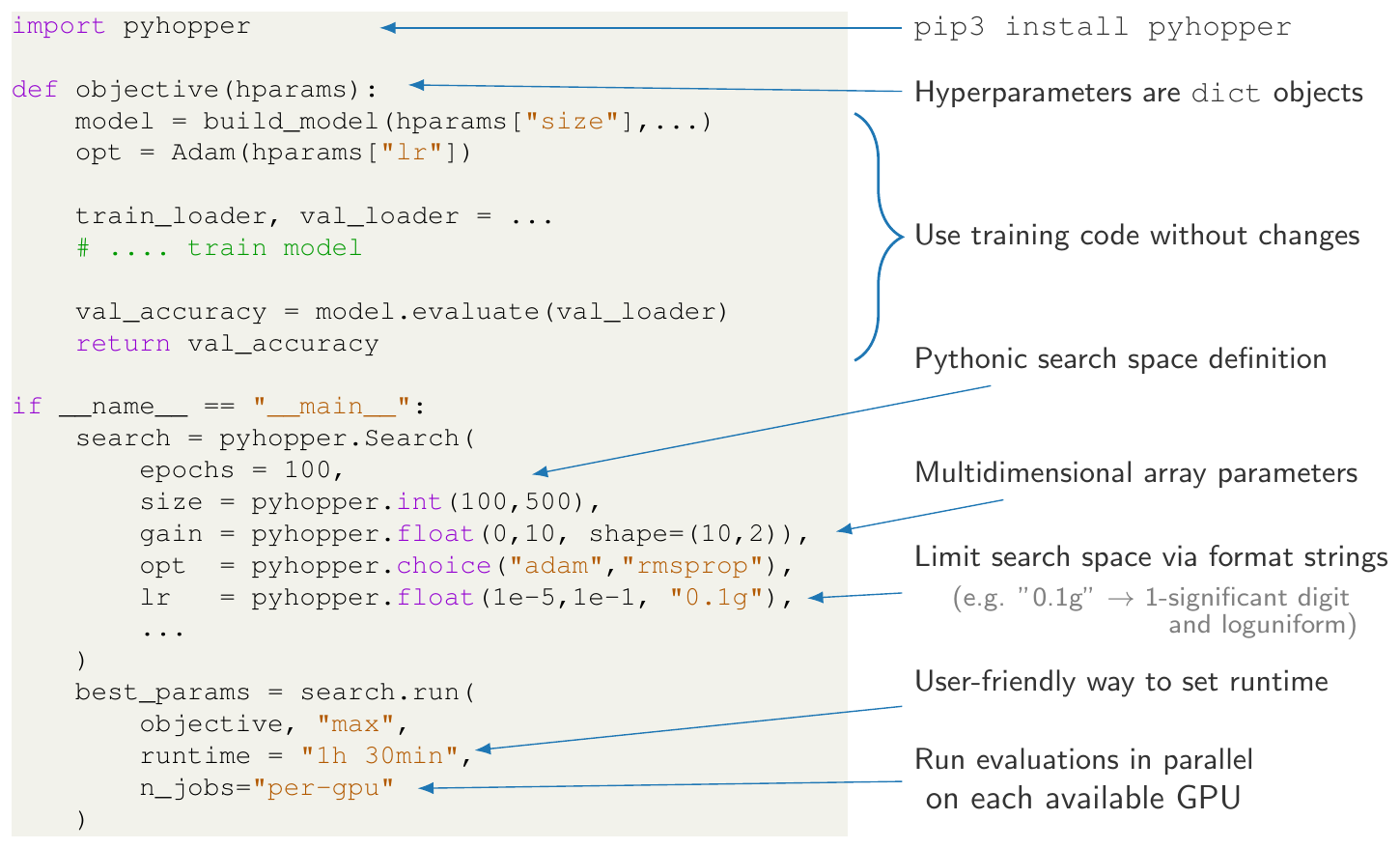}
    \caption{Visual abstract showing a typical use of PyHopper and some of its key features.}
    \label{fig:visabstract}
\end{figure}
\end{abstract}

\tableofcontents

\section{Introduction}

Modern machine learning (ML) research involves a considerable amount of hyperparameter tuning.
A hyperparameter is a value that is required to be set for training a machine learning model before the optimization process begins. For instance, the learning rate, i.e., the step size with which an optimization algorithm performs the next iteration towards minimizing a loss function, is a typical hyperparameter. Other examples of hyperparameters include the choice of the optimization algorithm, weight regularization factors, or simply the width and depth of a neural network. Hyperparameters are typically set before the optimization process and are not learned together with the model parameters. This makes finding the right set of hyperparameters challenging.

Changes in the hyperparameters drastically affect the performance of a trained ML model. For instance, a learning rate set too high or too low can make the difference between failing or solving a task. Moreover, the relation of the optimization process with respect to the hyperparameters is non-convex and non-differentiable. Consequently, the problem of finding the optimal hyperparameters could be formulated as a black-box optimization problem.

The dependency of ML models' performance on hyperparameters is a fundamental issue in machine learning research. This is because when we get to compare models with each other, ensuring a fair comparison between a new method and the previous baselines highly depends on the choice of hyperparameters. For example. suppose we want to compare the test performance of two different machine learning systems A and B. Shall we compare them with default hyperparameter settings, or shall we tune their hyperparameters, respectively? How much shall we tune them? How can we ensure we find the optimal settings for each model?

Moreover, consider the case where method A is already a well-established model whose optimal hyperparameter range is known. How much effort should we put into tuning the hyperparameters of B for a fair comparison?

The result of the challenges described above is that researchers can spend a tremendous amount of time tuning the hyperparameters of ML models. The pun grad-student-descent was coined describing how a typical graduate student working in machine learning spends most of their time manually tuning some hyperparameters.

Many algorithms and software packages for automatically tuning the hyperparameters have been proposed. Throughout, we will refer to these algorithms Hyperparameter Optimization (HPO) methods. However, the shape and size of hyperparameter optimization are very task-specific, and no one-fits-all solution or free lunch exists.

For instance, an HPO algorithm that performs well in low-dimensional problems may struggle to outperform simple baselines in higher-dimensional setups. Similarly, the smoothness and curvature of the objective surface can change drastically between two problem instances, making a one-fits-all solution impossible.
Moreover, the design of hyperparameter tuning packages often encounters contradictory specifications. For instance, an ideal package should be extensible, customizable, and rich in features, which, however, may steepen the learning curve and contradict our requirement that the package should be simple and easy to use.
 
In this work, we introduce Pyopper, a hyperparameter tuning platform tailored to the optimization frameworks we encounterin machine learning research (e.g., training neural networks). In particular, our HPO platform allows us to streamline the hyperparameter tuning procedures and scale to hundreds of tuning tasks with minimal effort. The key strengths of PyHopper are:
\begin{itemize}
\item An intuitive interface that integrates existing machine learning code requiring minimal changes
\item A highly customizable and robust optimization algorithm based on sequential Markov-chain Monte-Carlo sampling that scales to millions of hyperparameters
\item Numerous built-in utility methods to streamline common use cases, such as multi-GPU setup, checkpointing, and runtime scheduling.
\end{itemize}

\section{Related Works}

Numerous hyperparameter optimization algorithms have been proposed in the literature, each specialized for specific use cases and applications. Moreover, many publicly available HPO packages implement these algorithms. 
In this section, we first discuss the most important HPO algorithms and how they compare with each other. In the second part, we describe common HPO packages for Python and highlight their differences from PyHopper.

\subsection{Hyperparameter tuning algorithms}

\noindent \textbf{Grid Search} is arguably the most basic HPO. As its name suggests, Grid search spans a grid over the parameter space and evaluates every intersection point of the grid. The best configuration of the grid points is then returned as the best parameter. More advanced variations, such as iterative grid search, refine the grid resolution locally around the best-found parameter to explore the configuration space in more detail. The main advantage of Grid search is that it explores all parts of the configuration space. Thus it does not get easily trapped in local optima.
However, the major bottleneck of Grid search is that its complexity scales exponentially with the dimension of the configuration space, e.g., for eight hyperparameters, a grid with five ticks results in almost 400,000 intersection points that need to be evaluated.
Consequently, Grid search is only suitable for low dimensional configuration spaces, i.e., typically 2 or 3 hyperparameters.

\noindent\textbf{Sequential Model-Based Optimization} (SMBO)  \cite{hutter2011sequential} is a powerful black-box optimization paradigm. The key idea of SMBO is to fit a surrogate model to the already evaluated points to interpolate between unexplored parts of the configuration space. The surrogate model has a special structure that allows finding global optimums easily, e.g., analytically. These global optimums of the fitted surrogate model are then evaluated using the true objective function, and the observed objective values are used to update the surrogate model.

\noindent\textbf{Bayesian Optimization} (BO) extends SMBO by fitting distributions instead of deterministic functions to the evaluated points of the configuration space. The key benefit of BO over SMBO is that they allow modeling the uncertainty about the interpolated parts of the configuration space, i.e., the uncertainty increases the further away a point is from an already evaluated candidate configuration. Consequently, we can sample points in the configuration space that will maximize the gained information about the optimum of the black-box objective function.

The main advantage of SMBO and BO is that they become more and more accurate at finding optimal parameters the more information about the objective landscape becomes available. Moreover, BO allows tailoring the algorithm for specific applications by selecting the type of surrogate model used, i.e., adding prior knowledge about the optimization landscape.
However, this comes with the downside that SMBO and BO are less effective when little information (i.e., evaluated parameters) is available. Nonetheless, Bayesian optimization is often used in competition-winning toolkits due to its ability to add prior knowledge about the optimization landscape and adjust the algorithm to the particular type of the competition's optimization problem.
Gaussian processes (GPs) are a form of Bayesian optimization where multivariate normal distributions realize the surrogate model  \cite{bergstra2011algorithms}. 

\noindent \textbf{Tree-structured Parzen Estimator} (TPE) \cite{bergstra2011algorithms} is a sequential model-based optimization algorithm that can handle conditional configuration spaces efficiently. An example of such a conditional configuration would be the number of layers and corresponding hidden units in each layer of a neural network. Particularly, the number of layers in the fifth layer is only needed if the number of layers exceeds 4.

\noindent\textbf{Random Search} (RS) is another straightforward black-box optimization baseline. RS samples candidate solutions from a uniform distribution over the entire configuration space.
Despite its simplicity, RS can be competitive and outperform alternative algorithms in high-dimensional configuration spaces \cite{bergstra2012random}.
PyHopper's HPO algorithm starts with an RS to gain information about the objective surface and decide for the second phase which area of the configuration space to focus on.

\noindent\textbf{Markov chain Monte Carlo} (MCMC) is a family of methods for sampling from a probability distribution that cannot be described in a simple form explicitly but as the stationary distribution of a Markov chain.
MCMC can tackle black-box optimization problems by defining a stochastic process whose equilibrium distribution corresponds to the optima points of the objective function. 
The most fundamental MCMC optimization method takes the current best configuration and generates a new sample by adding random noise to it. If the objective function of the new sample is better than the current best configuration, we have a new best point, otherwise, the new sample is discarded. Such procedures are also referred to as \emph{local search} or \emph{hill climbing} in combinatorial optimization literature because they examine the neighborhood the current best solution for potentially even better candidates.

\noindent\textbf{Simulated Annealing} (SA) \cite{kirkpatrick1983optimization} further extends this idea by continuously decreasing the variance of the added noise, i.e., the "temperature", and also keeping worse new samples with a certain probability. SA has been shown to be very effective for approximately solving hard, non-differentiable optimization problems found in integrated circuit design and manufacturing.
The second phase of PyHopper's search algorithm realizes such an annealing-based local sampling strategy that gradually narrows down the searched area around the best found configuration. 

\subsection{Hyperparameter tuning packages}
There exist a large set of publicly available hyperparameter tuning packages. Here, we briefly discuss some of the most common tools and highlight some of their unique features.

\noindent\textbf{HyperOpt} \cite{bergstra2013making} is a hyperparameter tuning framework that provides an implementation of the Random Search and the Tree of Parzen Estimators optimization algorithms.
The specialty of HyperOpt is that parallelization is supported via Apache Spark or in a custom way through a database. This allows HyperOpt to integrate with an Apache Spark cluster easily at the additional cost of effort to set up and maintain the Apache Spark cluster. 

\noindent\textbf{Optuna} \cite{optuna2019} is a hyperparameter tuning framework developed by Preferred Networks. Optuna implements many common optimization algorithms and supports parallel evaluation through a MySQL database to which remote evaluation workers can connect. The main focus of Optuna is on experiment tracking and visualization of evaluated configurations.

\noindent\textbf{NeverGrad} \cite{nevergrad} is a black-box optimization library developed by Facebook Research. It implements various common gradient-free optimization algorithms and allows executing multiple configurations in parallel.

\noindent\textbf{keras-tuner} \cite{omalley2019kerastuner} is a hyperparameter tuning library building on top of the Keras API and Tensorflow 2.
The package implements common tuning algorithms, including Random Search, Bayesian optimization, and the HyperBand algorithm.
The major limitation of keras-tuner is that it does not support running multiple evaluations in parallel. 

\noindent\textbf{Autotune} \cite{koch2018autotune} is an HPO platform primarily focused on large-scale tuning of traditional machine learning models. Autotune implements several evolutionary sampling algorithm that can be combined during the search.

\noindent\textbf{Dragonfly} \cite{JMLR:v21:18-223} is a black-box optimization package that implements numerous variants of Bayesian Optimization algorithms. On-machine multiprocessing for parallel evaluation is available in Dragonfly. 

\noindent\textbf{Ray Tune}  \cite{liaw2018tune} is the hyperparameter library building on top of the Ray distributed computing framework.
Ray Tune provides an enormous set of different hyperparameter tuning algorithms. Moreover, Ray Tune can serve as a distributed evaluation engine for other hyperparameter tuning tools such as Optuna, Dragonfly, and Hyperopt. While Ray Tune is relatively flexible in terms of possible parallelization and tuning procedures, the large number of available algorithms can overwhelm the user by creating the meta-problem of finding the best HPO algorithm.

\section{PyHopper HPO Algorithm via Use Cases}
PyHopper's goal is to simplify and streamline the hyperparameter tuning processes of machine learning research tasks.
As the requirements for the hyperparameter optimization procedures can be relatively diverse, we focus on three typical HPO scenarios (i.e., use cases).
We first describe each use case when and where we may encounter them. Then, we explain the requirements and how PyHopper is designed to handle these scenarios.
\subsection{Use Case 1 - HPO with maximum resource usage}
Our first use case concerns the research on novel machine learning methods. This includes for instance, when a new training algorithm is developed, or a new deep learning architecture is tested against other baselines.
To test whether our new method can provide a significant improvement over existing approaches, we must evaluate it on a series of established benchmarks. 
Therefore, we must tune the method's hyperparameters to maximize the validation metrics on the given benchmark datasets, i.e., beating the other baselines.

The HPO budget in this setup is usually limited by the available compute resources. Hence, to maximize our HPO process, we must fully utilize our compute resources by carefully scheduling the tuning procedures.

PyHopper helps us maximize resource usage via two key features:
First, PyHopper expects the user to set the target runtime of the hyperparameter tuning process. For instance, this allows PyHopper to run overnight (or over the weekend) and finish the next day in the morning, thus fully utilizing our hardware during non-working hours. 

The second feature of PyHopper is continuously storing the tuning progress in checkpoint files. Therefore, we can safely run PyHopper on preemptive (i.e., spot) instances that cloud providers typically offer at a discount without risking any loss in data. When the spot instance becomes online again, PyHopper can continue its tuning process from where it left.

An expert machine learning engineer, typically have some empirical experience with machine learning models. Accordingly, they may want to include their expert views into the HPO initial setup. 
Ideally, we would like to use this implicit knowledge when running a more principled hyperparameter tuning algorithm. PyHopper allows adding the expert user's views on the optimal range of hyperparameter values to its queue. Moreover, PyHopper can interleave several phases of manual and automatic tuning.

\subsection{Use Case 2 - Fair comparison of multiple methods}
The second use case concerns research on a new dataset or a novel machine learning application. Here, we set out to assess how well different ML models perform on our new task. A key requirement in this experimental evaluation is to ensure fairness between the baseline models, i.e., no ML method is given an unfair advantage. 
For instance, not tuning the hyperparameters of each baseline method with a similar budget would introduce a hidden bias in our evaluation, as we might pick the optimal hyperparameters of one method but a suboptimal one for another.
Therefore, proper HPO is vital to ensure fairness when comparing different methods. 
PyHopper allows running fair evaluations by giving each method the same runtime budget for the tuning process. Alternatively, we can define the number of tuning steps or assign the same predefined sequence of hyperparameter candidates to be tested for each model.

Moreover, PyHopper records the history of the evaluated candidates and theircorresponding objective function. Hence we can draw statistical analysis on the dependency and sensitivity of each method on the hyperparameters, e.g., mean, standard deviation, or percentiles of the distribution.

\begin{algorithm}[t]
\caption{High-level description of PyHopper's MCMC sampling algorithm (maximization)}\label{alg:cap}
\begin{algorithmic}
\State \textbf{Input} Parameter space $\Omega$, objective function $f: \Omega \rightarrow \mathbb{R}$
\State $\theta_1,\dots \theta_k \gets $ random samples from $\Omega$ $\hfill \triangleright$ Random search (phase 1)
\State $\theta_{\text{best}} \gets \text{argmax}_{\theta_i}\{f(\theta_i\}$
\State temperate $\tau \gets 1$
\While{not timeout}
    \State $\theta \gets  \theta_{\text{best}}$ + random noise with temperature $\tau$  $\hfill \triangleright$ Local search (phase 2)
    \If{$f(\theta) > f(\theta_{\text{best}})$}
        \State $\theta_{\text{best}} \gets \theta$
    \EndIf
    \State decrease temperature $\tau$
\EndWhile
\State \Return $\theta_{\text{best}}$
\end{algorithmic}
\end{algorithm}

\begin{figure}[t]
    \centering
    \includegraphics[width=\textwidth]{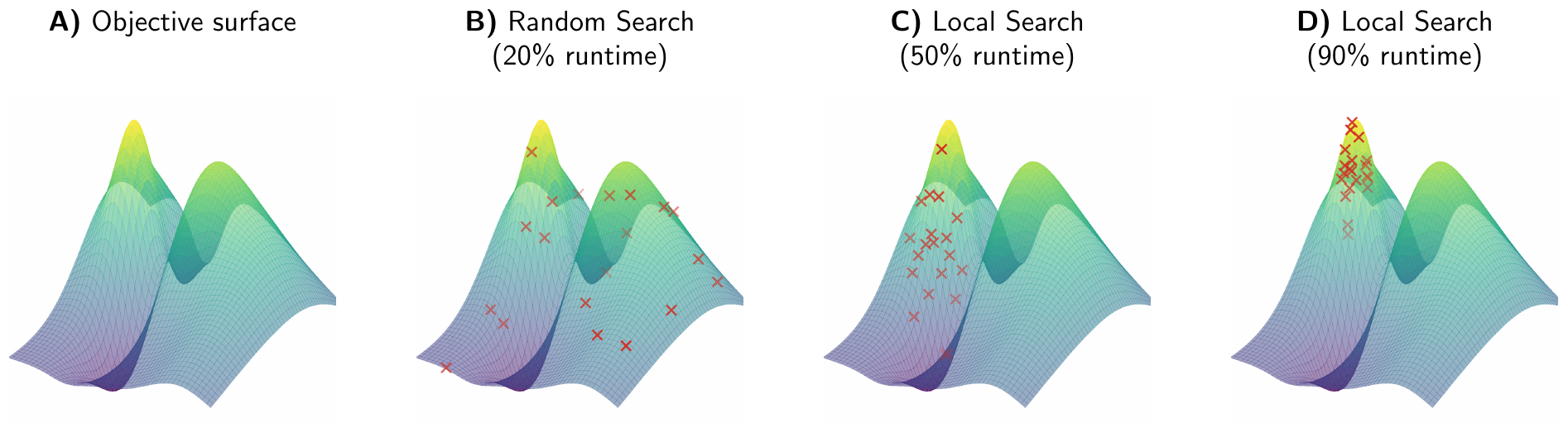}
    \caption{Example illustration of a 2-dimensional optimization problem and how PyHopper's optimization algorithm gradually narrows down the search area. \textbf{A)} Objective surface. \textbf{B)} Evaluated points during the Random Search (phase 1). \textbf{C)} Evaluated points during the beginning of the Local Search (phase 2). \textbf{D)} Evaluated points during the end of the Local Search (phase 2).}
    \label{fig:surface}
\end{figure}

\subsection{Use Case 3 - Black-box (gradient-free) optimization}
The third use case concerns general black-box (gradient-free) optimization problems found in research. 
For example, in robotics, we often have some unknown model or control parameters that we want to estimate and tune. 
For such applications, we can employ PyHopper with minimal effort. 
Firstly, PyHopper supports multi-dimensional NumPy arrays typed hyperparameters. Thus, we can tackle problems from one-dimensional parameters to search spaces with millions of dimensions. For instance, in the examples repository, we find PyHopper applied to gradient-free reinforcement learning by treating a neural network's weight matrices as hyperparameters.

Moreover, due to the simplicity of PyHopper's underlying optimization algorithm, we can define custom types and sampling strategies with as little as a single line of code. The \texttt{examples} repository includes the application of PyHopper to the traveling salesman problem (TSP) via a 2-top heuristic.

Although there exist toolkits specifically tailored for tackling these problems, the advantage of using PyHopper is that it allows rapid prototyping while taking care of boilerplate code, e.g., multiprocessing, scheduling, and the sampling algorithm.

\section{Optimization algorithm}

PyHopper revolves around a single optimization algorithm based on MCMC sampling.
The algorithm consists of two sequential phases, i.e., an exploration and an exploitation phase.
In phase 1, a Random Search draws random samples uniformly over the entire configuration space. The main idea of phase 1 is to gather information about the objective surface, i.e., which parts of the configuration space seem promising and which do not.

In the second phase, a sequential MCMC sampler takes the current best configuration and generates new samples by adding random noise.
The principle insight of phase 2 is to make incremental but consistent improvements to the current best solution.
To illustrate an example, phase 1 is about filtering out what hyperparameter combinations do not work at all, while phase 1 is about the details of the optimal ones, e.g., whether the best learning rate is 0.01 or 0.02.
A pseudocode representation of PyHopper's algorithm is shown in Algorithm \ref{alg:cap}.

\begin{wrapfigure}{r}{10cm}
\includegraphics[width=10cm]{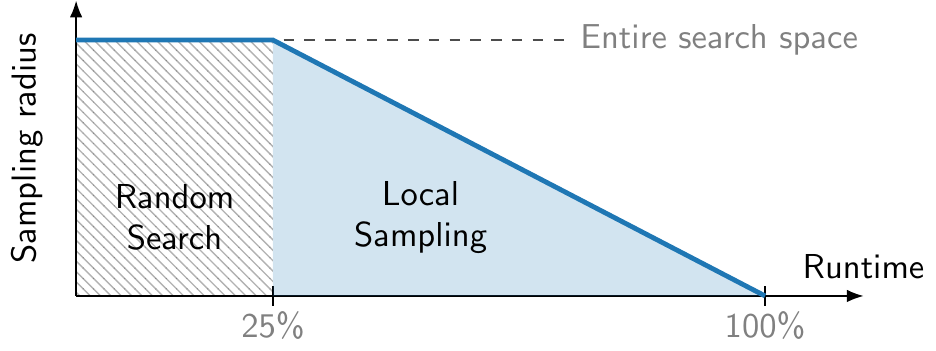}
\caption{Visualization of PyHopper's scheduling. At the beginning, a Random Search uniformly samples candidates from the entire configuration space. During the second phase, the Local sampling procedure gradually narrows the search area around the current best configuration. By default, the Random Search is scheduled to account for 25\% of the total runtime, while annealing of the Local sampling phase is scheduled for the remaining 75\%.}
\label{fig:schedule}
\end{wrapfigure}

PyHopper adopts the idea of simulated annealing and gradually decreases the magnitude of the noise over the runtime of the tuning process. Phase 2 is referred to as local sampling due to narrowing down the search area locally around the current best solution. 
The schematic of the two phases and corresponding scheduling is shown in Figure \ref{fig:schedule}. 
By default, PyHopper spends 25\% of the runtime performing a random search (phase 1). The "temperature," i.e., the noise variance, then linearly decreases over the remaining runtime in the second phase running the local search.

PyHopper requires the user to specify the target runtime of the tuning process, which provides two major benefits.
First, it allows exact scheduling between the two phases and the annealing process of phase 2. This ensures that the tuning process spends sufficient time exploring the configuration space and exploiting the promising areas.
The provided runtime comes with the additional benefit that we can schedule the tuning process to maximize the usage of available hardware. For instance, we can specify PyHopper to run overnight or over the weekend and be done by Monday morning. 
Figure \ref{fig:surface} visualizes an example of how a searchable area gradually focuses over the scheduled runtime.

The focus of PyHopper on a single optimization algorithm avoids the dilemma of having to decide between multiple tuning algorithms, i.e., the meta-problem of finding the best HP tuning algorithm, as well as it allows for streamlining the user interface.

PyHopper's algorithm is flexible and customizable. For example, we can skip phase 1 and directly let the local sampling algorithm improve on a set of hyperparameters the user provides.
Such scenarios often occur when the user finds some decently working hyperparameter through a manual search.
Moreover, PyHopper allows integrating custom sampling and local perturbation (i.e., mutation) strategies for special types of problems.
For instance,  the Travelling salesman problem (TSP) is an NP-complete combinatorial optimization problem that concerns finding the shortest roundtrip over a set of cities. With PyHopper, we can implement algorithms for heuristically solving the TSP with very minimal code.

\section{Parallelization}

PyHopper can parallelize the evaluations of hyperparameter candidates by executing the objective function on multiple processes simultaneously.
The sampling of new candidate solutions, scheduling, and invoking callback functions are all done in the main process. Thus, no code change is required by the user to use PyHopper's parallel executing engine.

\begin{figure}
    \centering
    \includegraphics{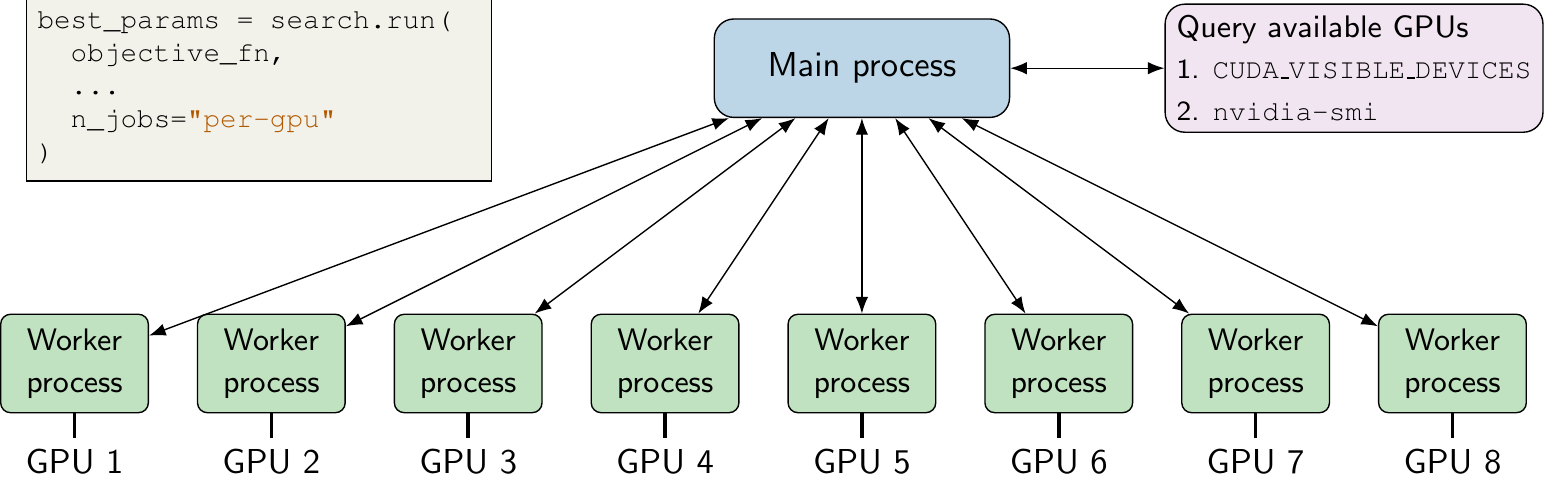}
    \caption{High-level description of the parallelization built-in by PyHopper. If the \texttt{n\_jobs} argument is set to \texttt{per-gpu}, the objective function is executed in parallel in subprocesses with only a single GPU visible to each process.}
    \label{fig:parallel}
\end{figure}

PyHopper is primarily intended to run on a machine with multiple NVIDIA GPUs installed (single-machine multi-GPU). We can spawn a parallel evaluation process for each available GPU with a single argument. PyHopper takes care of detecting the number of installed GPUs and setting the environment variables accordingly. 
Multi-node parallelization may be included in future versions of PyHopper.

\begin{figure}
    \centering
    \includegraphics[width=\textwidth]{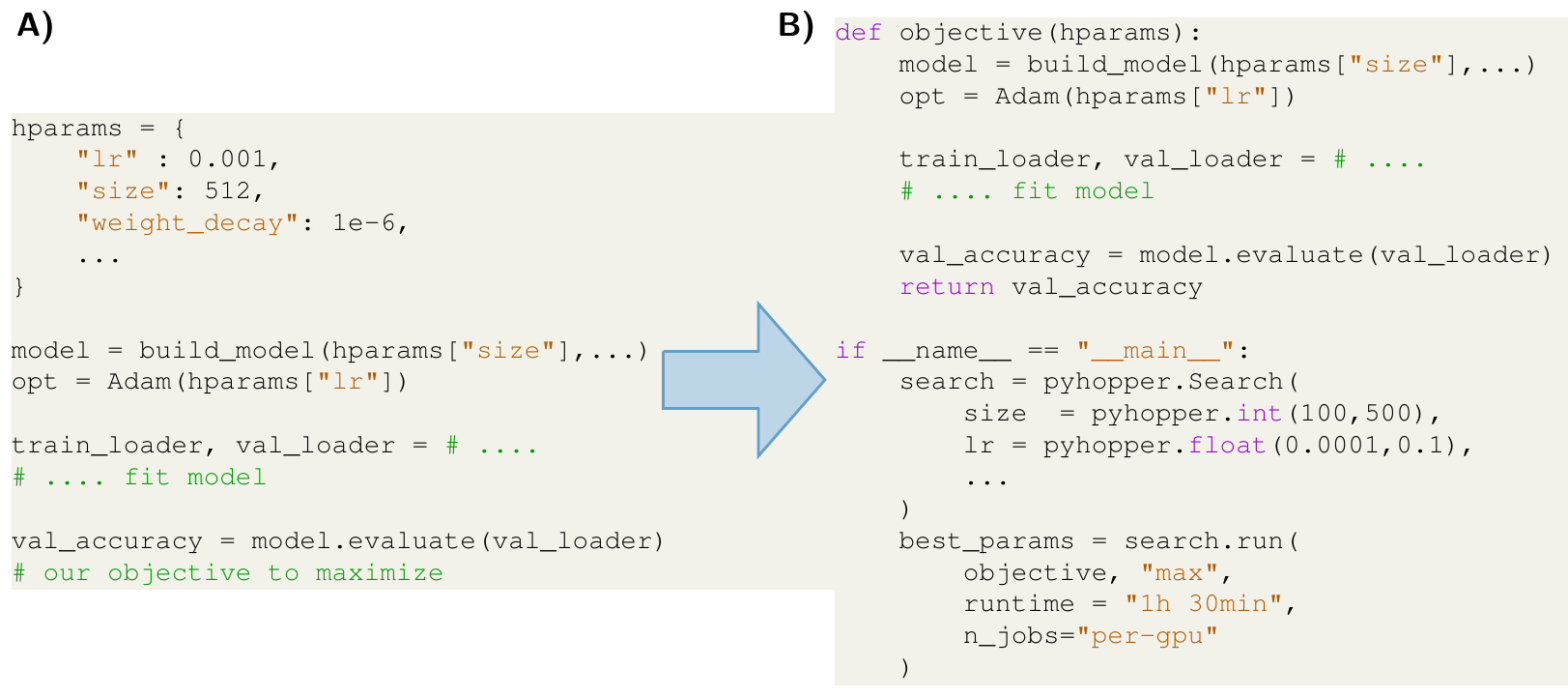}
    \caption{Example code snippet of how PyHopper integrates with existing ML code. \textbf{A)} Typical machine learning code for training a neural network. \textbf{B)} Adapted code for hyperparameter tuning with minimal required changes.}
    \label{fig:code-example}
\end{figure}

\section{API design}
Machine learning research can involve an enormous amount of hyperparameter tuning. 
PyHopper's API is designed to minimize the necessary changes that have to be made to the training pipeline and to simplify the integration of the tuned hyperparameter to other code.
Particularly, the user interface of PyHopper aims to remove the friction between the ML code and the hyperparameter tuning code. 

\begin{table}[]
    \centering
     \begin{tabular}{l|l}\toprule
    \textbf{Definition} & \textbf{Samples}\\\midrule
    \multicolumn{2}{l}{Integer parameters (uniform)}\\
    \ttx{  PyHopper.int(100,500)} & 350, 250, 500, ...\\
    \ttx{  PyHopper.int(100,500, multiple\_of=100)} & 400, 200, 100, ...\\
   \ttx{  PyHopper.int(0,10, shape=3)} & (5,2,7), (10,2,6), ...\\\midrule
   \multicolumn{2}{l}{Integer parameters (loguniform)}\\
    \ttx{  PyHopper.int(2,64, power\_of=2)} & 8, 4, 32, ...\\\midrule
    \multicolumn{2}{l}{Float parameters (uniform)}\\
    \ttx{  PyHopper.int(0,1)} & 0.5434, 0.83934, ...\\
    \ttx{  PyHopper.int(0,1, precision=1)} & 0.5, 1.0, 0.3, ...\\
   \ttx{  PyHopper.int(0,1, "0.1f")} & 0.5, 1.0, 0.3, ...\\
      \ttx{  PyHopper.int(-10,10, shape=2)} & (-6.22343, 1.5234), (0.3632,-8.90331), ...\\\midrule
      \multicolumn{2}{l}{Float parameters (loguniform)}\\
   \ttx{  PyHopper.int(1e-5,1e-3,log=True)} & 0.00023784, 0.000072342 ...\\
   \ttx{  PyHopper.int(1e-5,1e-3,log=True,precision=1)} & 2e-4, 5e-5, 8e-4, ...\\
  \ttx{  PyHopper.int(1e-5,1e-3,"0.1g")} & 2e-4, 5e-5, 8e-4, ...\\\midrule
   \multicolumn{2}{l}{Set parameters (unordered)}\\
  \ttx{  PyHopper.choice(["adam","sgd","rmsprop"]} & "sgd", "adam", ...\\\midrule
     \multicolumn{2}{l}{Set parameters (ordered)}\\
  \ttx{  PyHopper.choice([1,10,100],is\_ordinal=True)} & 10, 1, 100 , ...\\\bottomrule
    \end{tabular}
    \caption{List of supported datatypes in PyHopper and corresponding examples.}
    \label{tab:datatypes}
\end{table}

\subsection{Separation of concerns}
Our key idea is that any ML code, i.e., for both training and usage afterward, should be able to run without any dependency on the hyperparameter tuning package.
In particular, this means that the objective function, i.e., the training and validation, should contain any call to a function from the tuning package.
Consequently, PyHopper does not implement a \emph{define-by-run} API as done in other hyperparamter tuning packages \cite{optuna2019,omalley2019kerastuner}.
Instead, the interface between PyHopper and ML pipeline is represented by a Python dictionary (i.e., dict) object containing the hyperparameters.
Additionally, this design choice provides the advantage that the tuned hyperparameters in the form of a python dictionary can be easily stored, logged, and examined by the user.

The required adaptions of typical ML training pipelines to PyHopper are to wrap the training and validation into an objective function, define the configuration space, and run the search. 
The search space is set up by defining a template hyperparameter dictionary.
An example is shown in Figure \ref{fig:code-example}.

\subsection{Helpler functions}
PyHopper implements several helper and utility functions that simplify common tasks involved in hyperparameter tuning.
For instance, instead of defining the runtime of the tuning process in seconds, a string can be provided that will be parsed, supporting multiple units of time, e.g., hours or minutes.
Another example is the njobs argument, which, if set to "per-gpu" will take care of querying how many GPU devices are available on the system and scale the tuning process to run an evaluation on each device in parallel.

\subsection{Customization}
PyHopper allows defining of custom parameter types by defining custom sampling and mutation strategies. 
Additionally, PyHopper allows pruning candidates during the evaluation, e.g., if the validation accuracy does not reach a certain threshold in the first few epochs. 
Live feedback of evaluated candidates and current best configurations can be streamed through callback functionalities in PyHopper. For instance,  PyHopper provides a built-in checkpointing mechanism that continuously saves evaluated candidates and corresponding objective values in a file so that no information is lost if the machine crashes.

\subsection{Pruning algorithms}
The training of neural networks is an inherently stochastic process. For it instance, the random seed for the weight initialization affects the final accuracy of a model. This can be problematic for hyperparameter tuning, as we cannot fully trust the objective function always returns the same value. 
In particular, some evaluations might be lucky and report a bit higher performance in the objective function due to a particularly good random seed. 

Fixing the random seed in the training process avoids the stochasticity of the objective function. However, a fixed seed makes the tuning process potentially overfits the hyperparameters to the specific train-validation split. As a result, transferring the hyperparameters to a different train-validation split may result in a drop in performance.

A more reliable way to deal with this problem is to evaluate every HP configuration several times and report the mean. However, this significantly increases the computational cost of the tuning procedure. To counteract this cost explosion while maintaining the reliability of testing a configuration with several random seeds, we can employ pruning algorithms that stop the evaluation process of unpromising candidates already after their first evaluation.
In particular, if the first evaluation of an HP candidate indicates that this configuration has very little chance of being the best hyperparameter, the pruning algorithm will stop the remaining evaluations and discard the HP candidate.

The API design for the pruning interface was inspired by Optuna \cite{optuna2019}.

\section{Examples}
In this section, we provide the most fundamental examples demonstrating the real-world use of PyHopper.

\subsection{Available datatypes}
In Table \ref{tab:datatypes}, we list the available datatypes of hyperparameters in PyHopper. Moreover, Table \ref{tab:datatypes} demonstrates the most common customization patterns and corresponding samples.

\begin{figure}
    \centering
    \includegraphics[width=\linewidth]{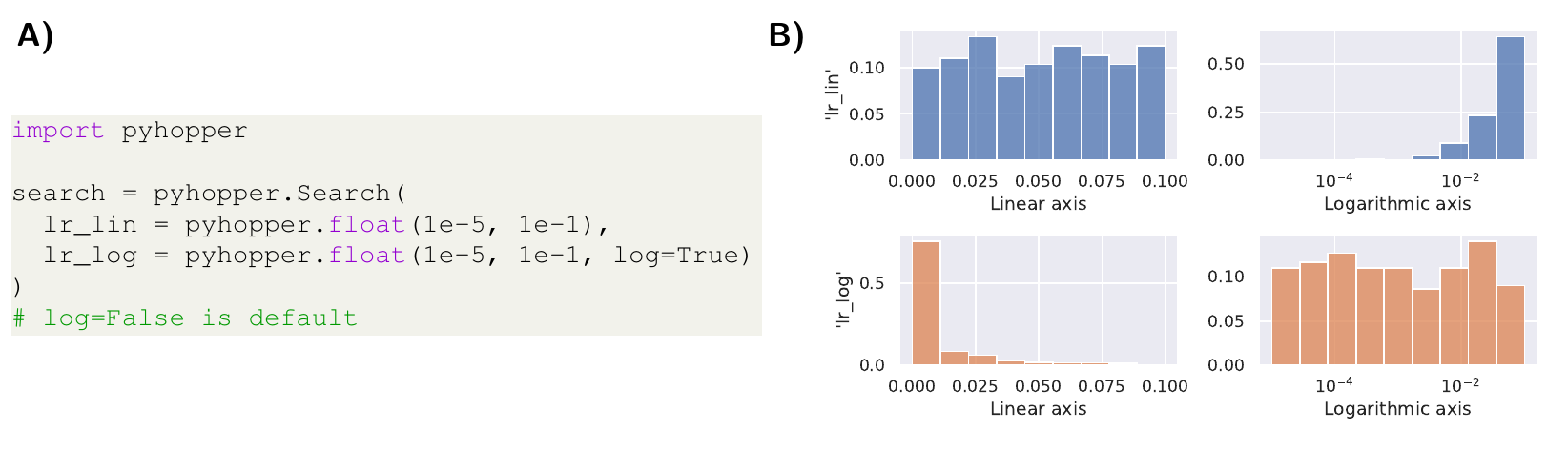}
    \caption{Difference between a linear (uniform) and logarithmically (loguniform) distributed float parameter. \textbf{A)} Logarithmical sampling of a float parameter can be enabled via the \texttt{log} argument. \textbf{B)} Resulting histogram of sampled values. The top row shows a uniformly and the bottom row a loguniformly distributed parameter. The two plots on the left use a linear x-axis, while the two plots on the right use a logarithmic x-axis. }
    \label{fig:linlog}
\end{figure}

For an efficient search, we distinguish between linearly and logarithmically distributed hyperparameters.
The default case is parameters with a linear search space and uniform density. 
This can be problematic for parameter spaces that range over several orders of magnitude. For instance, for a learning rate parameter spanning between $10^{-5}$ and $10^{-1}$, there is a 50\% chance that a uniform sample will be greater than $5\cdot 10^{-2}$, i.e., approximately the center of the interval $[10^{-5},  10^{-1}]$. Consequently, the optimization algorithm is biased toward sampling large values and might not explore values are the lower end of the spectrum with the same frequency.
A logarithmic sampling distribution, e.g., loguniform, resolves this issue and is therefore preferred for parameters that span multiple orders of magnitude.
In Figure \ref{fig:linlog}, we visualize sampling differences between a uniform (default) and loguniform distributed parameter. 

\subsection{Log-uniform and quantized distributions via format string} 
\begin{wrapfigure}{r}{10cm}
\includegraphics[width=10cm]{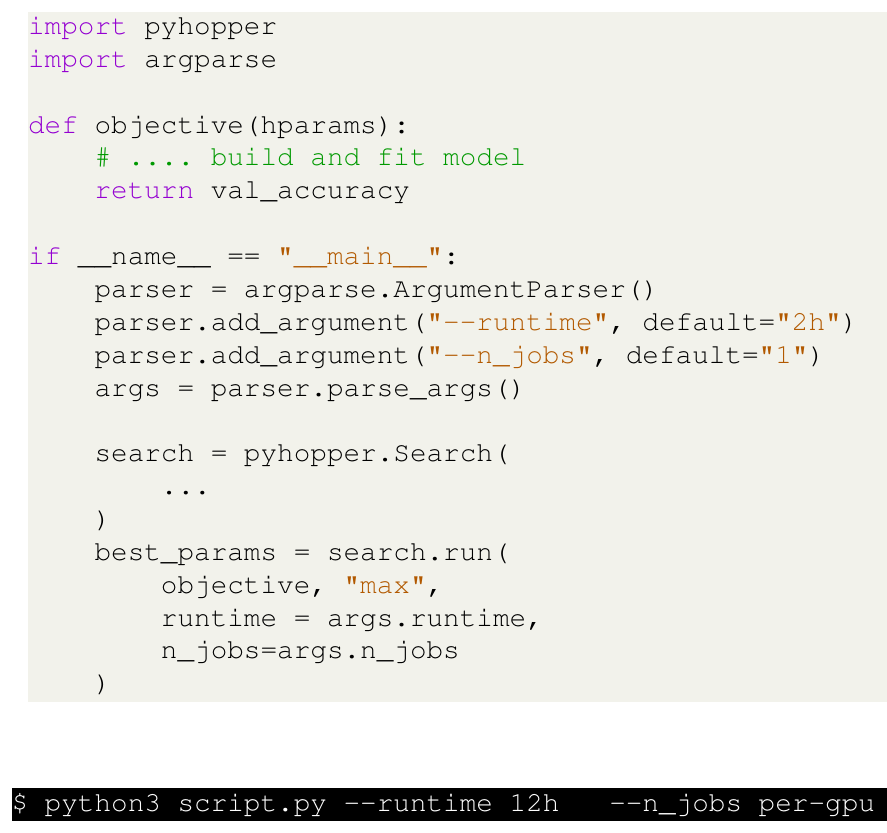}
\caption{Example of how PyHopper's run parameters are supposed to be forwarded from the command line arguments.}
\label{fig:argparse}
\end{wrapfigure}
Most Python developers are somewhat proficient with Python format strings of floating-point numbers. For instance the \texttt{":0.2f"} formats a floating-point number to two decimal digits after the comma, or \texttt{":0.1g"} to one significant digits in scientific notation (e.g., 3e-5). We re-purpose this concept to allow quantization and logarithmic sampling of the parameter space. For example, passing \texttt{":0.2f"} to a float-type PyHopper hyperparameter quantizes it to two decimal digits and uses a (linear) uniform search space. Contrarily, \texttt{":0.1g"} makes the float parameter sample from a log-uniform distribution quantized to one significant digit.

\subsection{Command line arguments}
PyHopper accepts its runtime and njobs argument in the form of a string that will be parsed. This allows directly forwarding command line arguments to PyHopper's run method and building an intuitive interface for the user with minimal code without limiting the freedom and flexibility of the developer.
An example of this command line argument forwarding is shown in Figure \ref{fig:argparse}.

\begin{figure}
    \centering
    \includegraphics[width=\linewidth]{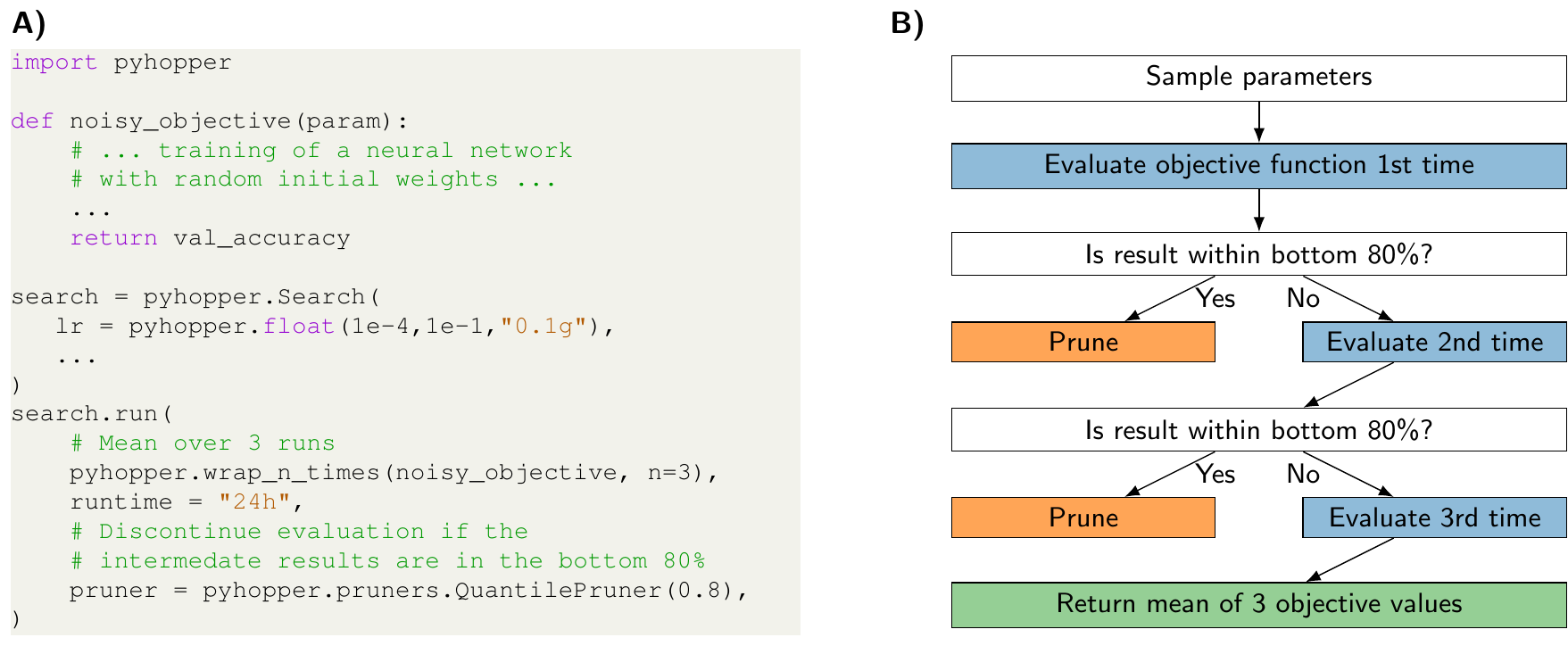}
    \caption{Example of how to deal with stochastic objective function and instantiate a pruning algorithm in PyHopper. \textbf{A)} Code snippet showing that only two lines of code are required to evaluate the mean of a noisy objective function and instantiate a pruning algorithm that prunes the candidate if the intermediate results are not within the top 20\%. \textbf{B)} Corresponding flow-chart involved in the pruning decision. }
    \label{fig:pruner}
\end{figure}

\subsection{Noisy objective and pruning}
As highlighted before, one way to deal with a stochastic objective function is to evaluate it several times and optimize the mean.
Moreover, we mentioned pruning algorithms to discontinue candidates that turn out to be unpromising already after the first evaluation.
In Figure \ref{fig:pruner}, we demonstrate that only two lines of code are required to implement this behavior. In particular, the \mbox{\texttt{PyHopper.wrap\_n\_times}} helper function wraps the objective function into a loop evaluating a given number of times. Moreover, the intermediate results are sent to the pruning algorithm, which decides whether to continue or prune the candidate.

\subsection{Fault tolerance and preemptive compute instances}
\begin{figure}
     \centering
     \begin{subfigure}[b]{0.35\textwidth}
         \centering
         \includegraphics[height=2.9cm]{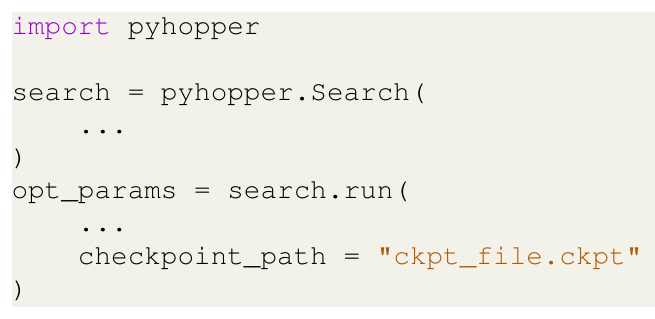}
         \caption{In case a filename is provided, PyHopper will resume the search from the checkpoint if the file exists. Continuous progress of the search will be stored in the file.}
         \label{fig:c1}
     \end{subfigure}
     \hfill
     \begin{subfigure}[b]{0.32\textwidth}
         \centering
         \includegraphics[height=2.9cm]{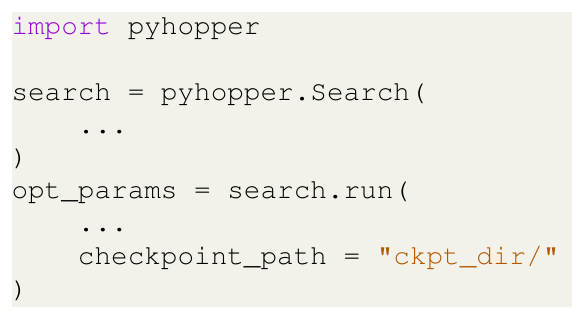}
         \caption{In case a directory is provided, PyHopper will create a new checkpoint within the directory. Progress is stored in the new checkpoint file.}
         \label{fig:c2}
     \end{subfigure}
     \hfill
     \begin{subfigure}[b]{0.25\textwidth}
         \centering
         \includegraphics[height=2.9cm]{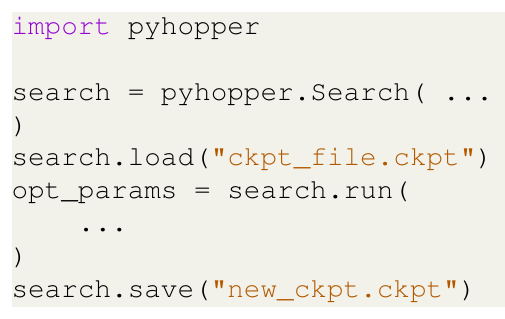}
         \caption{Manual checkpointing. Through the load and save functions, manual checkpoints can be created and loaded.}
         \label{fig:c3}
     \end{subfigure}
        \caption{Demonstration of a different way to use PyHoppers checkpointing mechanism}
        \label{fig:checkpointing}
\end{figure}

A typical hyperparameter tuning process can run for several days or even weeks. It is, therefore, necessary to prepare for unexpected events, such as power cuts or software bugs, to avoid a loss in data.
Additionally, to optimize costs, we could run the hyperparameter tuning process on preemptive cloud machines (i.e., spot instances), which can be shut down anytime.
PyHopper provides a checkpointing mechanism that continuously saves its internal state on the disk. Consequently, in case the tuning process is interrupted, e.g., by a power cut or spot instance shutdown, we can resume the tuning process from the last checkpoint. 
In Figure \ref{fig:checkpointing}, we demonstrate how to use the checkpointing mechanism of PyHopper.

\section{Experiments}

We set up an experimental evaluation to benchmark four popular hyperparameter optimization platforms: Optuna \cite{optuna2019}, HyperOpt\cite{bergstra2013hyperopt}, ray-tune\cite{liaw2018tune}, and PyHopper.
For a fair comparison, we define the exact same configuration spaces for all methods and allow each method to sample 30 hyperparameter configurations in total.
The objective function that should be optimized by the tools consists of training a Transformer model \cite{vaswani2017attention} on the IMDB sentiment analysis dataset \cite{imdb}. 
Our second experiment concerns the training of an LSTM network on the Walker2D kinematics modelling dataset \cite{lechner2020learning}.
The hyperparameters include, among others, the learning rate, number of attention heads, size of the LSTM cell \cite{hochreiter1997long}, and dropout rate applied to the word embedding. Random seed was fixed for the training (weight initialization and dataset shuffling).
The configuration space considered for both experiments are listed in Table \ref{tab:configspace} (IMDB task) and Table \ref{tab:configlstm} (Walker2D task) respectively.
The code for running the experiment is available at \url{https://github.com/pyhopper/comparison-examples}.

\begin{table}[t]
    \centering
        \caption{Configuration space of our experiment setup training a Transformer model \cite{vaswani2017attention} on the common IMDB sentiment dataset \cite{imdb}}
    \begin{tabular}{l|l}\toprule
         Hyperparameter & Range \\\midrule
         Learning rate & 1e-4 to 1-e2 (loguniform) \\
         Learning rate deacy & 0.2 to 1.0 (quantized in 0.1 steps)\\
         Warumup gradient steps & 100 to 1000 (quantized in 100 steps)\\
         Learning rate decay every n-th epoch & {5,10} \\
         Number of attention heads & 4 to 8 \\
         Dimension per attention head & 16 to 128 (quantized in 16 steps)\\
         Feedforward dimension & 64 to 512 (quantized in 64 steps)\\
         Weight decay & 1e-6 to 1e-4 (loguniform) \\
         Dropout rate & 0 to 0.3 (quantized in 0.1 steps)\\
         Number of layers  & 2 to 6 \\
         Apply LayerNorm on word embedding & True, False \\
         Word embedding dropout rate & 0 to 0.3 (quantized in 0.1 steps)\\\bottomrule
    \end{tabular}
    \label{tab:configspace}
\end{table}

\begin{table}[]
    \centering
     \caption{Configuration space of our experiment setup training an LSTM network \cite{hochreiter1997long} on the Walker2D kinematics modelling dataset \cite{lechner2020learning}}
    \begin{tabular}{l|l}\toprule
         Hyperparameter & Range \\\midrule
          LSTM cell size & 64 to 512 (quantized in 64 steps)\\
         Learning rate & 1e-4 to 1-e2 (loguniform) \\
         Learning rate deacy & 0.2 to 1.0 (quantized in 0.1 steps)\\
         Warumup gradient steps & 100 to 500 (quantized in 100 steps)\\
         Learning rate decay every n-th epoch & {5,10,20,50} \\
         Weight decay & 1e-6 to 1e-4 (loguniform) \\\bottomrule
    \end{tabular}
    \label{tab:configlstm}
\end{table}

Both ray-tune and PyHopper provide 1-line multi-GPU parallelization capabilities, which we enable for the evaluation. This significantly reduces the runtime on our machines (8 Titan RTX GPUs for the IMDB task, and 2 A6000 GPUs for the Walker2D task), however, it potentially limits the optimization as strictly sequential testing of hyperparameters results in the maximum information being available for generating informed new candidates.

The results in Table \ref{tab:results} show that Optuna, ray-tune, and Pyhopper could find competitive performing hyperparameter settings for both tasks. 
Moreover, the runtimes indicate that Pyhopper concluded its search process the fastest. In particular, for IMDB, PyHopper's runtime is more than 10x faster than Optuna and HyperOpt, and 1.2x better than ray-tune. Moreover, we observed that HyperOpt was not able to find good hyperparameter candidates. We hypothesize that HyperOpt's Bayesian optimization engine did not allocate the 30 available samples to properly cover the search space. 

\begin{table}[t]
    \centering
    \caption{Results of our experimental comparison of common hyperparameter tuning packages. The objective function was set to maximize the validation accuracy and minimize the validation mean-squared error (MSE) for the IMDB and Walker2D task respectively. For the tools supporting 1-line parallel multi-GPU execution (ray-tune and Pyhopper), parallel execution was enabled. Optuna \cite{optuna2019}, ray-tune \cite{liaw2018tune}, and Pyhopper demonstrate competitive optimization performance.}
    \begin{tabular}{l|cc|cc}\toprule
        Dataset & \multicolumn{2}{c}{IMDB \cite{imdb}} & \multicolumn{2}{c}{Walker2D \cite{lechner2020learning} }\\\midrule
         HPO & Best validation & Runtime & Best validation & Runtime  \\
        platform & accuracy & (minute) & MSE & (minutes) \\\midrule
         Optuna \cite{optuna2019} & 87.47\% & 823 & \textbf{0.863} & 42\\
         HyperOpt \cite{bergstra2013hyperopt} & 50.00 \%  & 808 & 1.956 & 42 \\
         ray-tune \cite{liaw2018tune} & 87.46\% & 104 & 0.897 & 23 \\
         PyHopper (ours) & \textbf{87.98}\% & \textbf{80} & 0.878 & \textbf{20}  \\\bottomrule
    \end{tabular}
    \label{tab:results}
    \vspace{-5mm}
\end{table}

\section{Limitations}
There cannot be a perfect hyperparameter tuning package, as some features of what makes a good HP tuner might be contradictory. For instance, implementing several different optimization algorithms might be both an advantage and a disadvantage.
Instead, each hyperparameter tuning package comes with tradeoffs that were made for specific application areas in mind.
The main tradeoff for PyHopper is the focus on a single optimization algorithm. As highlighted before, having a single algorithm provides the advantage the user does not need to bother with which algorithm to run, i.e., the meta-problem finding the best hyperparameter tuning algorithms that finds the best hyperparameters. 
Moreover, the focus on the single MCMC sampler allows the user to easily customize the algorithm as there are fewer layers of abstraction than packages that implement several optimization routines.
Nonetheless,  PyHopper's optimization algorithm comes with a set of disadvantages. 
First, Bayesian optimization-based approaches might be preferable for lower dimensional problems and for problems where many candidate configurations can be evaluated.
For such problem instances, enough information about the objective surfaces is available for a model-based algorithm to accurately approximate the black-box objective function via a surrogate model and overcome the problem of getting stuck in local optima.
Moreover, running multiple hyperparameter algorithms in parallel might be feasible for problems where our compute budget is large enough.
Consequently, packages that can try several different algorithms might have an advantage over PyHopper in such cases.

\FloatBarrier
\section{Conclusion}
PyHopper is a customizable, open-source, and plug-and-play hyperparameter optimization engine, that can be integrated with advanced training jobs with minimal effort and low cost, generating competitive models compared to existing well-established packages.

\bibliographystyle{plain} 
\bibliography{references}

\end{document}